\documentclass{article}

\usepackage[preprint]{corl_2026} % Use this for the initial submission.

\usepackage{amsfonts}
\usepackage{amsmath}
\usepackage{amssymb}
\usepackage{amsthm}
\usepackage{graphicx}
\usepackage{enumitem}
\usepackage{booktabs}

\usepackage{wrapfig}

\usepackage[table]{xcolor}

\title{Neural Navigation Functions for Zero-Shot Generalizable Motion Planning}
\author{
Benjamin D. Shaffer$^{1}$ \quad
Pei-An Hsieh$^{2}$ \quad
Brooks Kinch$^{1}$ \quad
Nathaniel Trask$^{1}$ \quad
M. Ani Hsieh$^{1}$ \\
University of Pennsylvania, United States \\
$^{1}$Department of Mechanical Engineering and Applied Mechanics \\
$^{2}$Department of Electrical and Systems Engineering \\
\texttt{\{ben31,mya\}@seas.upenn.edu}
}
\begin{document}
\maketitle

%===============================================================================

\begin{abstract}
We introduce Neural Navigation Functions (Neural-NF), a learned reactive navigation function capable of zero-shot transfer across unseen environment geometries.
Neural-NF places data-driven adaptation within a structured elliptic planner, where the navigation objective is learned while planner structure is preserved by construction. Specifically, intrinsic Laplacian-derived features are mapped to local PDE coefficients, and solving the resulting boundary value problem produces a globally consistent value function on each target domain. For every admissible learned model, the resulting policy is collision-free, provides monotonic descent and a global minimum at the goal by construction. This admits a linearly-solvable optimal-control interpretation for any parameter setting. Empirically, Neural-NF achieves strong zero-shot transfer across diverse geometries and outperforms learned planners that directly predict the value function by up to a $5\times$ improvement.  
\end{abstract}

\section{Introduction}
Robot motion planning enables autonomous navigation and manipulation in complex environments, requiring methods that are adaptable to task objectives, robust under changes in geometry, and reliable throughout the entire operation.
Artificial potential functions are widely used as they simultaneously solve the path planning and controller synthesis problems within a single framework.
% used for robot motion planning
By defining a scalar field over the configuration space that encodes goal locations as global minima, the negative gradient directly yields a smooth feedback control policy that drives the robot to the goal while avoiding obstacles,
producing closed-loop feedback policies that can be evaluated in real time without explicit trajectory search \citep{khatib1986real}. However, classical potential field constructions are known to suffer spurious local minima.  
Navigation-function and PDE-based feedback planners address this shortcoming by providing formal guarantees of convergence to the goal with no spurious attractors, and therefore provide a compelling representation for reliable motion planning \citep{rimon1990exact}. Their behavior, however, is specified through a prescribed objective.

Learning-based motion planners \citep{matada2024generalizable, ni2022ntfields, ni2025physics} offer the complementary ability to adapt navigation behavior from data, but direct prediction of paths, policies, or value fields does not preserve navigation-function structure by construction and must generalize the resulting global planning map beyond the training geometries. 
There is a growing interest in learning-based alternatives that can preserve the desirable properties of navigation functions while generalizing across diverse and novel environments. 

To this end, we propose Neural Navigation Function (Neural-NF), a learning-based \emph{local-to-global} planning  framework for robot motion in which the navigation function is represented as the solution of a \emph{learnable elliptic PDE}.  A trained model provides spatially varying operator coefficients from intrinsic geometric features of the environment and the value function is recovered by solving the resulting linear boundary value problem on the target domain. 
\begin{wrapfigure}{r}{0.55\textwidth}
    \vspace{-\intextsep}
    \centering
    \includegraphics[width=\linewidth]{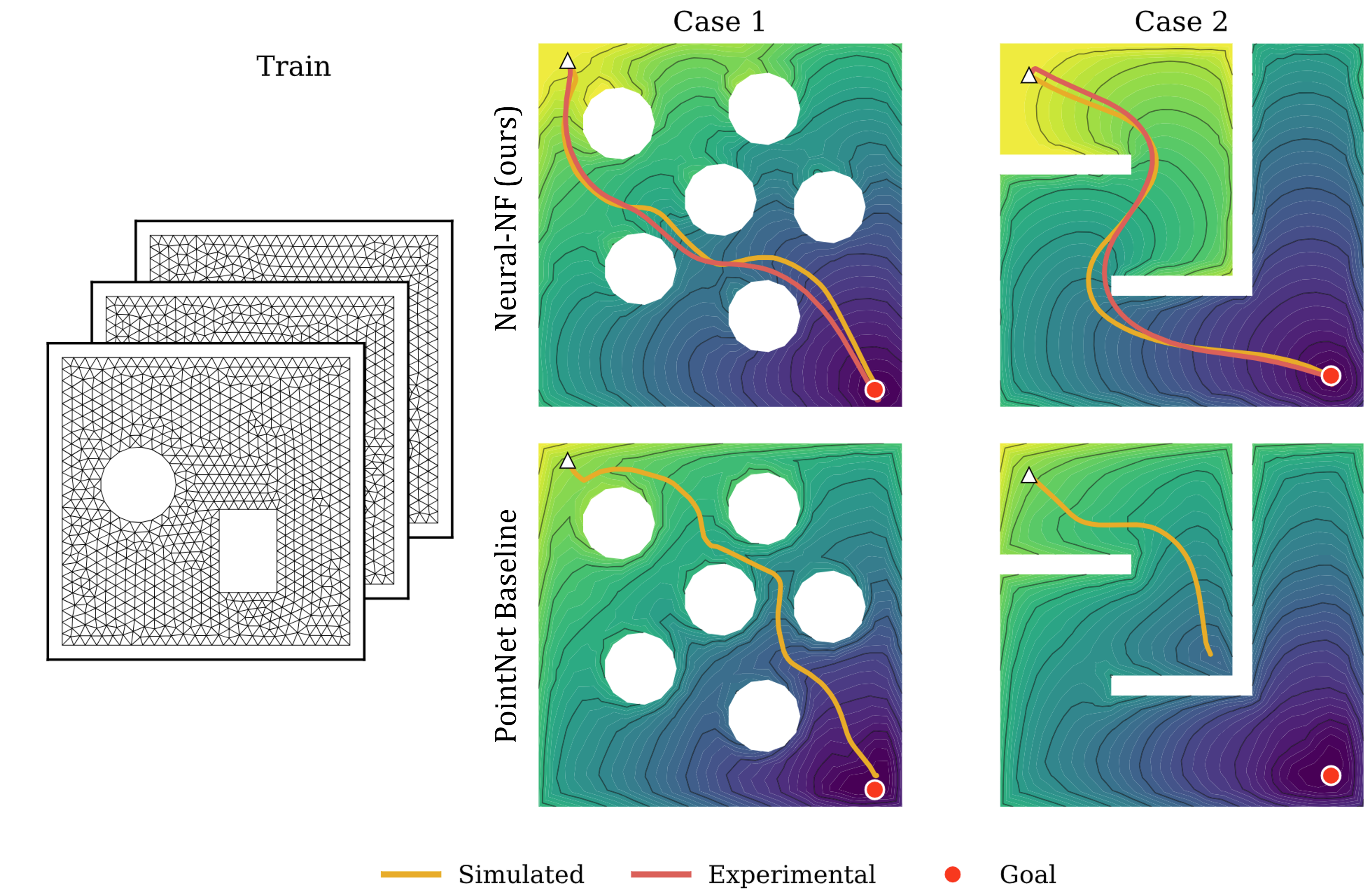}
    \caption{Neural-NF produces accurate value functions on unseen, complex, out-of-training-distribution geometries while preserving structural guarantees on non-boundary-collision, monotonicity, regularity and enabling flexibility to objective functions via learning. \textbf{Case 1} extrapolates the number of obstacles in the domain while \textbf{Case 2} introduces a qualitatively different feature in the interior corners, where the baseline learned planner fails.}
    \label{fig:fig1}
\end{wrapfigure}
Learning therefore determines the navigation problem being solved, while global policy construction, boundary compatibility, regularity, and monotone value descent remain properties of the planner construction for every admissible learned coefficient field.

The coefficients are trained end-to-end by differentiating through the linear solve, allowing adaptation within this planning class to objectives such as travel time, clearance, and spatial running costs. The resulting elliptic formulation also admits a linearly-solvable stochastic optimal-control interpretation, in which the learned coefficients determine the running cost and control metric~\cite{kappen2005linear, todorov2009compositionality, dvijotham2012linearly}. Empirically, Neural-NF enables zero-shot transfer to geometry distributions not seen during training, including from as few as a single training geometry, and substantially outperforms direct learned planners while requiring less training data.
Our contributions are:

\begin{enumerate}[leftmargin=8pt,itemsep=0pt,parsep=2pt,topsep=0pt]
\item \textbf{Zero-shot geometry transfer.} We introduce a learnable planner that transfers across unseen geometries, including transfer to qualitatively different geometry classes.

\item \textbf{Data-efficient training.} Neural-NF achieves strong transfer from as few as a single training geometry by learning reusable local operator coefficients.

\item \textbf{Structured planning properties.} For every admissible parameter setting, the learned PDE defines a well-posed planner with guaranteed global maximum principle, collision free dynamics, an optimal-control interpretation, and monotone descent of the value function.

\item \textbf{End-to-end adaptable objectives.} The navigation coefficient models are trained by differentiating through the finite element solve, allowing optimization for general navigation objectives.
\end{enumerate}

\section{Related Work}

\paragraph{Feedback motion planning.}
Global feedback planners represent navigation through a scalar field over the free-space domain whose gradient induces a closed-loop policy. Navigation functions establish convergence and obstacle-avoidance properties under appropriate geometric assumptions~\cite{koditschek1990robot}, while potential-field and PDE-based planners construct related navigation fields through boundary value problems posed on the environment geometry~\cite{wang2000new, conner2003composition, gingras2010path, masoud2012motion, palm2014fluid}. Hamilton--Jacobi methods formulate planning through value functions associated with system dynamics and path costs~\cite{takei2010practical}. Linearly-solvable stochastic control gives a closely related construction in which a logarithmic transformation maps a class of optimal-control value functions to linear elliptic desirability equations~\cite{todorov2009compositionality, dvijotham2012linearly, theodorou2010generalized}.

\paragraph{Learned motion planning.}
Learning-based motion planning includes methods that predict planning actions, paths, or value representations from environment and goal information~\cite{wang2021survey, qureshi2019motion, tamar2016value, chen2025online}. Continuous-field methods more directly target feedback planning by learning environment fields or time-to-goal representations over the domain~\cite{li2021learning, ni2022ntfields}. A substantial recent line of physics-informed neural planners introduces eikonal structure, geometric constraints, domain decomposition, and task-specific extensions into these learned fields~\cite{ni2023progressive, ni2024physics, ni2025physics, ren2025physics, liu2025physics, shen2024pc, li2025koopmotion}. Operator-learning approaches extend global value-field prediction across environments and goal specifications~\cite{matada2024generalizable}, while recent viscous value representations incorporate second-order Hamilton--Jacobi structure into learned navigation values~\cite{viswanath2026physics}. Learned components have also been used within graph-search and sampling-based planning procedures to improve planning efficiency while retaining structure from the underlying algorithm~\cite{ichter2018learning, yonetani2021path, hassidof2025train}. \citet{rousseas2021harmonic, rousseas2024reactive} use a similar learnable potential field navigation approach but within a reinforcement learning framework. 

\paragraph{Structured learning.}
Differentiable optimization layers, deep equilibrium models, and implicit-differentiation frameworks provide tools for training through converged optimization or fixed-point problems~\citep{amos2017optnet,bai2019deep,blondel2022efficient}. Physics-informed and PDE-constrained learning methods similarly train models through differential structure, including work on learning conductivity and solving eikonal equations with neural networks~\citep{tartakovsky2020physics,bin2021pinneik}.
Finite element exterior calculus and related structure-preserving discretizations expose the geometric operators used in these implicit models~\citep{arnold2018finite,trask2022enforcing,actor2024data}. Recent work has used these ideas for generalizable scientific machine learning and for robotics- in adaptive sensing~\citep{kinch2025structure,shaffer2026structure,shaffer2026meshfree,shaffer2025multi,shaffer2025physics}. \citet{actor2024data} in particular presents the learned Hodge star formulation which is most similar to our implementation of the learned PDE.
We adapt this framework for feedback motion planning.

\section{Neural Navigation Functions (Neural-NF)}

\subsection{Problem statement}
\label{sec:problem_statement}

Let $g \in \mathcal{G}$ denote a geometric specification of the robot workspace ${\cal W}$, {\it i.e.}, obstacle and workspace boundary positions and geometries.  Let $\Omega_g \subset \mathbb{R}^d$ ($d \in \{2,3\}$) denote the free-space with boundary $\partial\Omega_g$ and assume a holonomic robot whose position is given by $x \in \Omega_{g}$.  Given a goal configuration $G \subset \Omega_g$, we assume robot kinematics given by 
\begin{equation}
    \dot{x}(t) = u(x(t); g, G), \qquad x(0) = x_0 \in \Omega_g,
\label{eq:dynamics}
\end{equation}
and seek a stationary feedback policy $u(x(t); g,G)$ that drives the robot to $G$ while remaining in $\Omega_g$.
In the navigation function formulation \cite{koditschek1990robot}, $u$ is defined as the negative gradient of a scalar navigation function $V$. 
We aim to learn a mapping $(g, G) \mapsto V$ that produces effective navigation policies across geometries, enabling \emph{generalization} to unseen instances and \emph{zero-shot transfer} across geometry classes.
Specifically, we assume geometries are drawn from an underlying distribution $\mathcal{P}$ and train on a finite set $\mathcal{G}_{\rm{train}}=\{g_i\}_{i=1}^{N_{\rm{train}}}$, with $g_i \sim \mathcal{P}$.
The learned planner is evaluated on unseen geometries $g \in \mathcal{G}_\mathrm{eval}$ drawn from $\mathcal{P}$ (generalization), as well as on geometries drawn from a distinct distribution $g \sim \hat{\mathcal{P}}$ (zero-shot transfer).

Broadly, we contrast learning-based approaches which seek to directly learn a mapping $(g,G)\mapsto V$ via a trainable parameterization $\theta$, and classical planning approaches which define $V$ satisfying $\mathcal{A}(V;g,G)=0$ where $\mathcal{A}$ is a pre-defined planning optimality condition. We combine these approaches by learning local navigation costs which are globally assembled through a structured PDE solve, preserving both the flexibility of learned approaches and the rigorous guarantees of traditional methods. Concretely, we learn $(g,G)\mapsto \mathcal{A}_\theta$ and then recover $V$ by solving $\mathcal{A}_\theta(V;g,G)=0$ via an intermediary desirability variable \eqref{eq:desire_pde} described below.

\subsection{Learned elliptic value function and induced policy}
\label{sec:method_description}

Elliptic PDE-based planners represent navigation as the gradient of a scalar value function defined implicitly as the solution of a boundary value problem on $\Omega_g$. This class of methods provides strong guarantees across geometries: (i) monotonic descent on value function, (ii) global maximum principles, and (iii) exact treatment of boundary conditions (e.g. no-collision).
We extend this framework to a learnable setting, defining a geometry and objective dependent family of feedback motion planners within the structured class. The motion planner is parameterized as the solution of a PDE with spatially varying coefficients defined by intrinsic geometric descriptors, yielding a flexible model that adapts across environments while preserving these guarantees.

\begin{equation}
    \label{eq:pipeline}
    \underbrace{\Omega_g}_{{\text{domain}}}
    \;\xrightarrow{\;\text{\S\ref{sec:intrinsic_features}}\;}\;
    \underbrace{\Phi_g}_{\text{intrinsic features}}
    \;\xrightarrow{\;\text{\S\ref{sec:method_description}}\;}\;
    \underbrace{\mathcal{Q}_\theta:(k_\theta, c_\theta)}_{\text{local navigation model}}
    \;\xrightarrow{\;{\rm{global}\atop\rm{solve}}\;}\;
    \underbrace{V}_{\text{navigation function}}
    \;\xrightarrow{\;eq.\; \ref{eq:desire_transform}\;}\;
    \underbrace{u}_{\text{policy}}
\end{equation}

For a given input domain $\Omega_g \subset \mathbb{R}^d$, with discrete representation as a simplicial mesh with nodal positions $X=\{x_i\}_{i=1}^N$, we construct local intrinsic features $\Phi_g \in \mathbb{R}^{N\times r}$ from Laplacian-derived quantities on this mesh, see Section~\ref{sec:intrinsic_features}.

A shared pointwise neural field maps the input features to two nodal coefficient fields:
\[
(k_\theta, c_\theta) = \mathcal{Q}_\theta(\Phi_g),\qquad k_\theta,c_\theta \in \mathbb{R}^N
\]
where $k_\theta > 0$ is the scalar conductivity and \(c_\theta >0\) is a running cost, with positivity enforced by construction (via softplus activation).
We then define the diagonal conductivity matrix
\(
K_\theta(x) = k_\theta(x)\, I.
\)

Given $K_\theta,\;c_\theta$, we solve for the desirability $z_\theta$ via the elliptic problem
\begin{equation}
    \label{eq:desire_pde}
    \nabla\cdot(K_\theta \nabla z_\theta) - c_\theta z_\theta = -\rho_{G}
    \quad \text{in } \Omega_g, 
    \qquad (K_\theta \nabla z_\theta)\cdot n = 0 \ \text{on } \partial\Omega_g,
\end{equation}
discretized on the mesh, which defines the system $\mathcal{A}_\theta$ from the categorization in Section~\ref{sec:problem_statement}. Here $\rho_{G}$ is a localized source encoding the goal region, for the optimal control equivalence this can instead be enforced via boundary conditions (Section~\ref{sec:lsoc}).
The goal-forced formulation is used throughout for experimental convenience; the structural properties are preserved in both cases.
The navigation value function and policy are
\begin{equation}
    \label{eq:desire_transform}
    V_\theta = -\log z_\theta,\qquad
    u_\theta = K_\theta \nabla \log z_\theta.
\end{equation}

\begin{wrapfigure}{r}{0.5\textwidth}
    \vspace{-\intextsep}
    \centering
    \includegraphics[width=\linewidth]{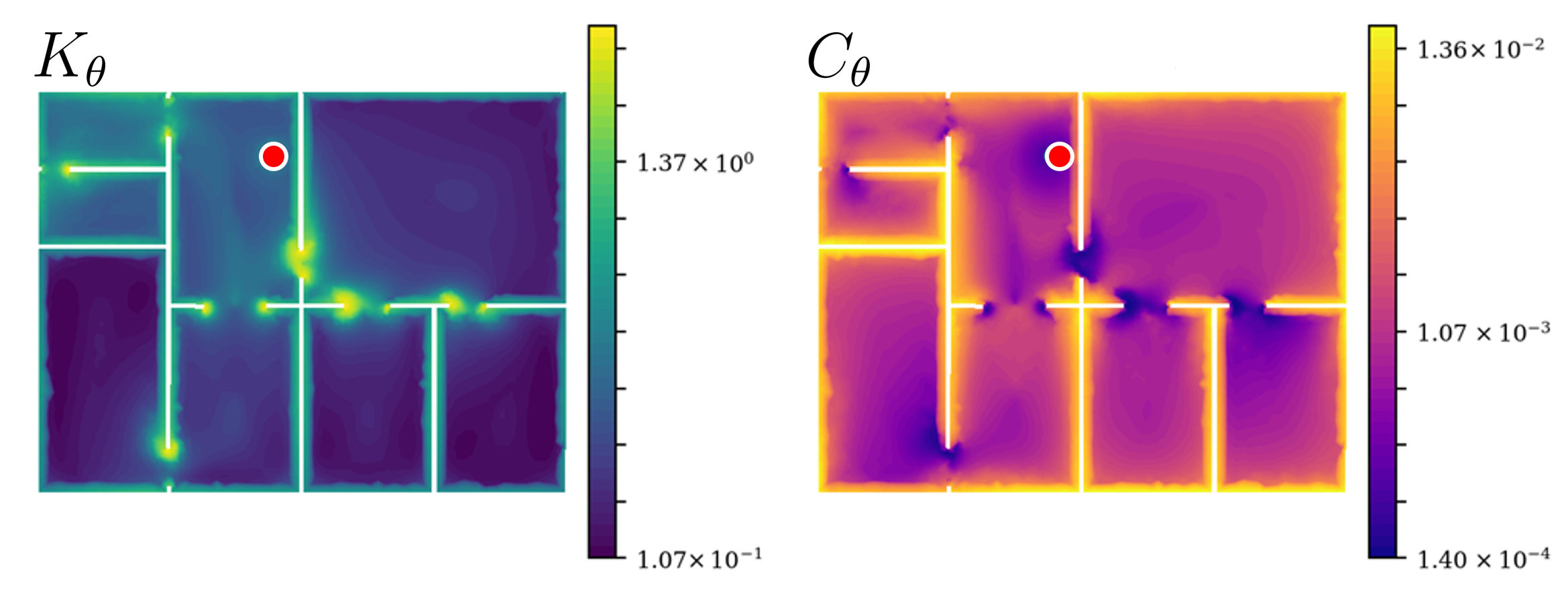}
    \caption{The model output of (left) the learned conductivity, $K_\theta$, and (right) the learned running cost, $C_\theta$, demonstrate the local, geometric nature of the learned terms. Particularly in the conductivity, where narrow corridors correspond to high conductivity and open spaces result in low conductivity. The goal location is shown in red.}
    \label{fig:learned_conductivity_vertical}
\end{wrapfigure}

This construction defines a local mapping $\Phi_g(x) \mapsto (K_\theta(x), c_\theta(x))$, so learning operates pointwise in the geometry, while global consistency is enforced by the PDE solve, thereby separating local adaptation from global structure and ensuring consistent navigation fields.
In practice, training gradients ($\partial_\theta\mathcal{L}(u)$, for loss function $\mathcal{L}$) are obtained by differentiating through the discretized linear solve of \eqref{eq:discrete_desirability}.

\noindent\emph{Remark.} Under standard conditions (bounded $K_\theta$ with $\alpha>0$, bounded \(c_\theta   >0\), and Lipschitz $\Omega_g$), the problem admits a unique positive solution $z_\theta$, ensuring the policy is well-defined.

\subsection{Discretization by Finite Element Exterior Calculus}
\label{sec:discretization}

We discretize \eqref{eq:desire_pde} using the lowest-order finite element exterior calculus (FEEC) complex~\cite{arnold2018finite}. This framework has recently been adopted in machine learning as it provides a flexible and general basis for constructing structure preserving discretizations in terms of differential forms \cite{kinch2025structure}. Let \(z\in\mathbb{R}^{N_0}\) denote the nodal desirability values and let \(d_0:\mathbb{R}^{N_0}\to\mathbb{R}^{N_1}\) be the incidence matrix mapping node values to oriented edge differences (the topological gradient), where $N_0$ and $N_1$ denote the number of nodes and edges of the mesh. We define \(M_0\) and \(M_1\) as the Whitney \(0\) and \(1\)-form mass matrices.
The discrete problem is then,
\begin{equation}
    \left(
    d_0^\top M_1(k_\theta)d_0 + M_0(c_\theta)
    \right)z_\theta=b_G,
    \label{eq:discrete_desirability}
\end{equation}
where \(M_1(k_\theta)\) and \(M_0(c_\theta)\) are the conductivity and
running-cost weighted Hodge operators and \(b_G\) encodes the goal forcing. 
Using the diagonal discrete representations of \(K_\theta\) and \(C_\theta\), with nodal conductivity interpolated to edges
\(
    (K_\theta)_{(i,j),(i,j)}
    =
    \frac{1}{2}\left(k_{\theta,i}+k_{\theta,j}\right)
\)
for mesh edges $(i,j)$,
we define \(M_1(k_\theta):=K_\theta^{1/2}M_1K_\theta^{1/2}\), and define \(M_0(c_\theta)\) analogously, see \citep{actor2024data}.
The homogeneous Neumann condition is natural (automatically enforced) in this weak formulation, ensuring non-collision with domain boundaries. Positivity of \(K_\theta\) and positivity of \(C_\theta\) give the standard coercivity conditions for a well-posed discrete solve.

This form exposes the local-to-global structure used by the model. The geometry-dependent operators \(d_0\), \(M_0\), and \(M_1\) are assembled once for a mesh, while learning only changes the sparse coefficient fields \(K_\theta\) and \(C_\theta\). Therefore a new goal or coefficient realization can be evaluated by reusing the same sparse geometric operators, without reassembling a weighted finite element Laplacian, as would be needed in a conventional differentiable FEM implementation.

\subsection{Intrinsic geometric features}
\label{sec:intrinsic_features}

To enable transfer across geometries, the input representation should encode intrinsic shape independent of coordinates and discretization while capturing both local and global structure. Operators derived from the Laplacian provide such a representation and are widely used in geometric learning as stable, multi-scale descriptors of shape~\cite{sharp2022diffusionnet, sun2009concise}.

We use four feature channels: heat kernel signatures \(\Phi_{\mathrm{HKS}}\), which encode multi-scale local shape over $T$ channels, Poisson depth \(\Phi_{\mathrm{Poisson}}\), which encodes boundary and obstacle proximity, a goal indicator \(\Phi_{\mathrm{goal}}\), which represents the target location via a Poisson solve with a point source on the goal, and an alignment feature \(\Phi_{\mathrm{align}}\), which compares goal direction with the boundary-depth field. These channels provide a coordinate-free representation of the geometry and goal-conditioned task, while leaving global coupling to the elliptic solve. Full definitions are given in Appendix~\ref{app:intrinsic_features}.
The full feature map is the concatenation
\[
\Phi_g = [\,\Phi_{\mathrm{HKS}},\;\Phi_{\mathrm{Poisson}},\;\Phi_{\mathrm{goal}},\;\Phi_{\mathrm{align}}\,] \in \mathbb{R}^{N\times(T+3)},
\]
which is used as the input to the coefficient networks. While not a unique choice, we show empirically in Section~\ref{sec:results} that these provide a strong basis for generalizable learning.

\subsection{Model training}

The proposed model defines a differentiable mapping from geometry and goal $(g, G)$ to a control field via the solution of an elliptic PDE. This enables multiple training paradigms depending on available supervision. In particular, the model can be trained (i) from trajectory-level objectives via rollouts (e.g., reinforcement learning) \citep{rousseas2021harmonic}, (ii) by enforcing PDE- or control-based constraints (e.g., PINN-style residual minimization such as eikonal losses) \cite{matada2024generalizable}, or (iii) via direct supervision of the value function or control policy when available.

In this work, we adopt a supervised formulation. Given a dataset $\mathcal{D} = \{(\Omega_g, G, V^{*})\}_i^{N_{\mathrm{train}}}$ with reference value functions $V^{*}$, we optimize
\(\min_\theta \; \mathbb{E}_{\mathcal{D}} \big[ \| V_\theta - V^{*} \|^2 \big],\)
where $V_\theta$ is obtained by solving the PDE defined by the learned coefficients. Gradients with respect to $\theta$ are computed by differentiating through the linear solve.

\section{Planner properties}

% \subsection{Navigation properties}

The structured learning approach imposes planner-level constraints independently of the learned parameters. First, the homogeneous Neumann condition gives \((K_\theta\nabla z_\theta)\cdot n=0\) on \(\partial\Omega_g\), so the induced continuous policy is tangent to obstacle boundaries. Further, in the interior, trajectories following \( \dot x = u_\theta(x) \) monotonically decrease the navigation value function:
\[
\frac{d}{dt}V_\theta(x(t))
= - \nabla \log z_\theta(x)^\top K_\theta(x)\nabla \log z_\theta(x) \le 0.
\]
Therefore \(V_\theta\) is a Lyapunov function for the closed-loop dynamics, with equality only at critical points of \(z_\theta\). 
Further, the elliptic problem class ensures that \(V_\theta\) has no spurious interior minima outside the goal region and attains its global minimum within \(G\), see Sec. 6.4 of \citet{evans2022partial}.
To demonstrate the impossibility of non-goal interior minima, consider a local minimum of \(V_\theta\), which is also local maximum of \(z_\theta\). At any
interior local maximum \(x\notin G\), we have
\(\nabla z_\theta(x)=0\), so we require
\(
\nabla\cdot(K_\theta\nabla z_\theta)(x) \leq 0.
\)
Since \(\rho_G\) is only supported in \(G\), the eq. \ref{eq:desire_pde} gives
\[
\nabla\cdot(K_\theta\nabla z_\theta)(x) = c_\theta(x)z_\theta(x)>0 \qquad \text{for } x\notin G,
\]
therefore there cannot be interior local minima outside $G$.

As in classical navigation-function constructions, non-goal critical points may remain.
We show in Appendix~\ref{app:geometric_interpretation} that this planning class corresponds precisely to cost-to-go navigation objectives under viscous regularization, which is a general and useful objective class, corresponding to the simplest second-order formulation.

\paragraph{Optimal-control interpretation}
\label{sec:lsoc}
For the Dirichlet-goal formulation, the learned desirability problem admits an exact linearly-solvable stochastic optimal control interpretation \citep{kappen2005linear, todorov2009compositionality}. Specifically, $z_\theta$ is the desirability function of a reflected controlled diffusion whose diffusion tensor and control metric are both determined by $K_\theta$, and whose state-dependent running cost is determined by $c_\theta$. Under the logarithmic transform $V_\theta = -\log z_\theta$, the optimal stationary feedback
coincides with the Neural-NF policy in \eqref{eq:desire_transform}. Thus, learning $(K_\theta, c_\theta)$ can be interpreted as learning a structured family of stochastic navigation objectives. The full statement and derivation are provided in Appendix~\ref{app:lsoc}.

\paragraph{Geometric inductive biases}
\label{sec:geometric_biases}
The local-to-global construction imposes geometric structure prior to training, restricting the learned planner to a hypothesis class whose outputs transform consistently with the geometry and whose learned component is local and reusable.

\emph{Equivariance.} Since features are built from the mesh Laplacian without a fixed coordinate frame, the coefficient fields and $V_\theta$ transform by pullback under rigid motions, giving an equivariant policy: $u_\theta(Rx;Rg,RG)=R\,u_\theta(x;g,G)$ for any rotation $R$ (formal statement in Appendix~\ref{app:equivariance}).

\emph{Locality and Stability.}
Geometry enters Neural-NF only through intrinsic Laplacian-derived features and the elliptic solve. The Laplacian-based features such as HKS vary smoothly under regular geometric deformation, and standard elliptic boundary value problems vary continuously under regular perturbations of their domains and coefficients~\cite{sun2009concise,delfour2011shapes}.
As a result, globally dissimilar environments sharing local features (corridors, corners) will result in similar learned coefficients, while global consistency is assembled by the PDE solve. Planner quality therefore depends only on the quality of local coefficient prediction and discretization, provided the local feature distribution remains similar, which motivates the strong geometric transfer demonstrated empirically.

\begin{figure}
    \centering
    \includegraphics[width=0.9\linewidth]{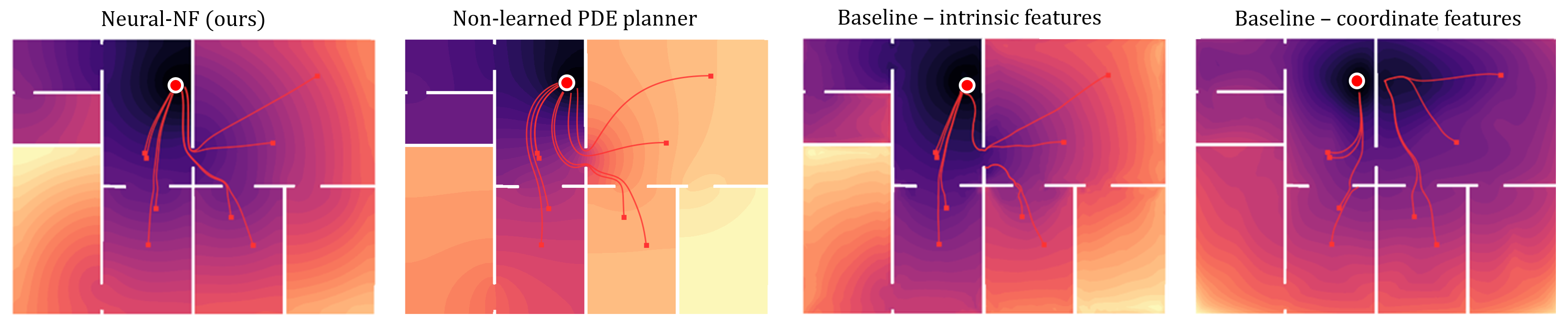}
    \caption{Neural-NF accurately reproduces the desired value function on an unseen geometry while producing a regular navigation policy. The unlearned PDE provides global policy structure but poor value reconstruction, whereas direct prediction from intrinsic features improves scalar accuracy without yielding a reliable gradient policy. Neural-NF combines transferable intrinsic features with global elliptic structure to produce accurate and navigable potentials.}
    \label{fig:sample_paths}
\end{figure}

\section{Results}
\label{sec:results}

\begin{table}[h]
\small
\centering
\caption{Comparison of $L_2$ error on the learned value function on test geometries for each of the considered datasets. We evaluate on both unseen test geometries drawn from the same distribution as training (ID) and from a different distribution (OOD). In both cases and for all experiments, Neural-NF results in substantial improvements over direct-prediction baselines; this is particularly pronounced for more complex geometries (e.g. \emph{Maze}) and for the OOD comparisons.}
\label{tab:main_table}
\begin{tabular}{lrrrrr}
\toprule
\textbf{ID} $\| V-V_{\textrm{target}}\|_2\, (\%)$ & Square & Disk & Maze & House Expo & City Streets \\
\midrule
PointNet &  \cellcolor{red!10} 2.8 $\pm$ 0.9 & 17.9 $\pm$ 13.5 & 40.5 $\pm$ 10.8 & 21.3 $\pm$ 12.4 & 11.4 $\pm$ 7.1 \\
GNN & 3.0 $\pm$ 1.4 & 16.7 $\pm$ 11.3 & 39.8 $\pm$ 9.6 & 19.6 $\pm$ 11.1 & 10.2 $\pm$ 6.3 \\
DeepONet & 4.2 $\pm$ 1.8 & 18.2 $\pm$ 13.2 & 40.2 $\pm$ 9.7 & 21.2 $\pm$ 11.9 & 14.4 $\pm$ 9.7 \\
GNO & 3.4 $\pm$ 1.5 &  17.3 $\pm$ 12.0 &  38.6 $\pm$ 10.2 & 19.6 $\pm$ 10.5 & 10.1 $\pm$ 5.9 \\
MGN & 3.0 $\pm$ 1.5 & 16.3 $\pm$ 10.2 & 39.4 $\pm$ 9.8 &  19.1 $\pm$ 10.4 &  9.4 $\pm$ 5.5 \\
Intrinsic & 4.7 $\pm$ 1.7 & \cellcolor{red!10} 7.0 $\pm$ 3.2 & \cellcolor{red!10} 4.5 $\pm$ 1.7 & \cellcolor{red!10} 7.8 $\pm$ 4.0 & \cellcolor{red!10} 6.6 $\pm$ 1.9 \\
\textbf{Neural-NF} & \cellcolor{blue!5} \textbf{1.5} $\pm$ 0.5 & \cellcolor{blue!5} \textbf{3.0} $\pm$ 3.7 & \cellcolor{blue!5} \textbf{3.7} $\pm$ 3.8 & \cellcolor{blue!5} \textbf{5.6} $\pm$ 7.9 & \cellcolor{blue!5} \textbf{6.5} $\pm$ 5.8 \\
\midrule
\textbf{OOD} $\| V-V_{\textrm{target}}\|_2\, (\%)$ & Square & Disk & Maze & House Expo & City Streets \\
\midrule
PointNet &  \cellcolor{red!10} 3.9 $\pm$ 1.7 & 20.1 $\pm$ 13.7 & 44.0 $\pm$ 9.7 &  23.9 $\pm$ 8.4 & 24.8 $\pm$ 11.8 \\
GNN & 4.4 $\pm$ 2.1 & 18.3 $\pm$ 11.8 & 41.8 $\pm$ 8.3 & 28.2 $\pm$ 10.9 & 23.7 $\pm$ 12.1 \\
DeepONet & 5.3 $\pm$ 2.2 & 23.2 $\pm$ 15.7 & 41.8 $\pm$ 8.5 & 25.4 $\pm$ 10.6 & 54.2 $\pm$ 15.8 \\
GNO & 4.3 $\pm$ 1.8 & 18.9 $\pm$ 12.5 & 42.2 $\pm$ 8.7 & 17.0 $\pm$ 4.2 & 23.6 $\pm$ 12.4 \\
MGN & 3.9 $\pm$ 1.8 & 18.7 $\pm$ 11.2 & 45.3 $\pm$ 8.6 & 17.1 $\pm$ 5.3 & 23.2 $\pm$ 12.2 \\
Intrinsic &  6.0 $\pm$ 2.0 & \cellcolor{red!10} 7.9 $\pm$ 3.9 & \cellcolor{red!10} 16.3 $\pm$ 8.9 & \cellcolor{red!10} 9.4 $\pm$ 2.3 & \cellcolor{blue!5} \textbf{10.8} $\pm$ 5.6 \\
\textbf{Neural-NF} & \cellcolor{blue!5} \textbf{2.1} $\pm$ 1.1 & \cellcolor{blue!5} \textbf{2.8} $\pm$ 3.2 & \cellcolor{blue!5} \textbf{7.4} $\pm$ 5.1 & \cellcolor{blue!5} \textbf{8.1} $\pm$ 3.0 &  \cellcolor{red!10} 11.5 $\pm$ 10.3 \\
\bottomrule
\end{tabular}
\end{table}

We evaluate on mesh-based 2D navigation domains constructed from synthetic environments and canonical navigation datasets, with separate splits for in-distribution test geometries and out-of-distribution geometry classes. For each geometry, we sample multiple goal locations and train using supervised targets \(V_{\mathrm{target}}\) computed via FMM for the eikonal objective. All methods are trained on the same geometry/goal splits and evaluated using normalized value function error \(\|V_\theta - V_{\mathrm{target}}\|_2\).
\begin{wrapfigure}{r}{0.4\textwidth}
    \vspace{-\intextsep}
    \centering
    \includegraphics[width=\linewidth]{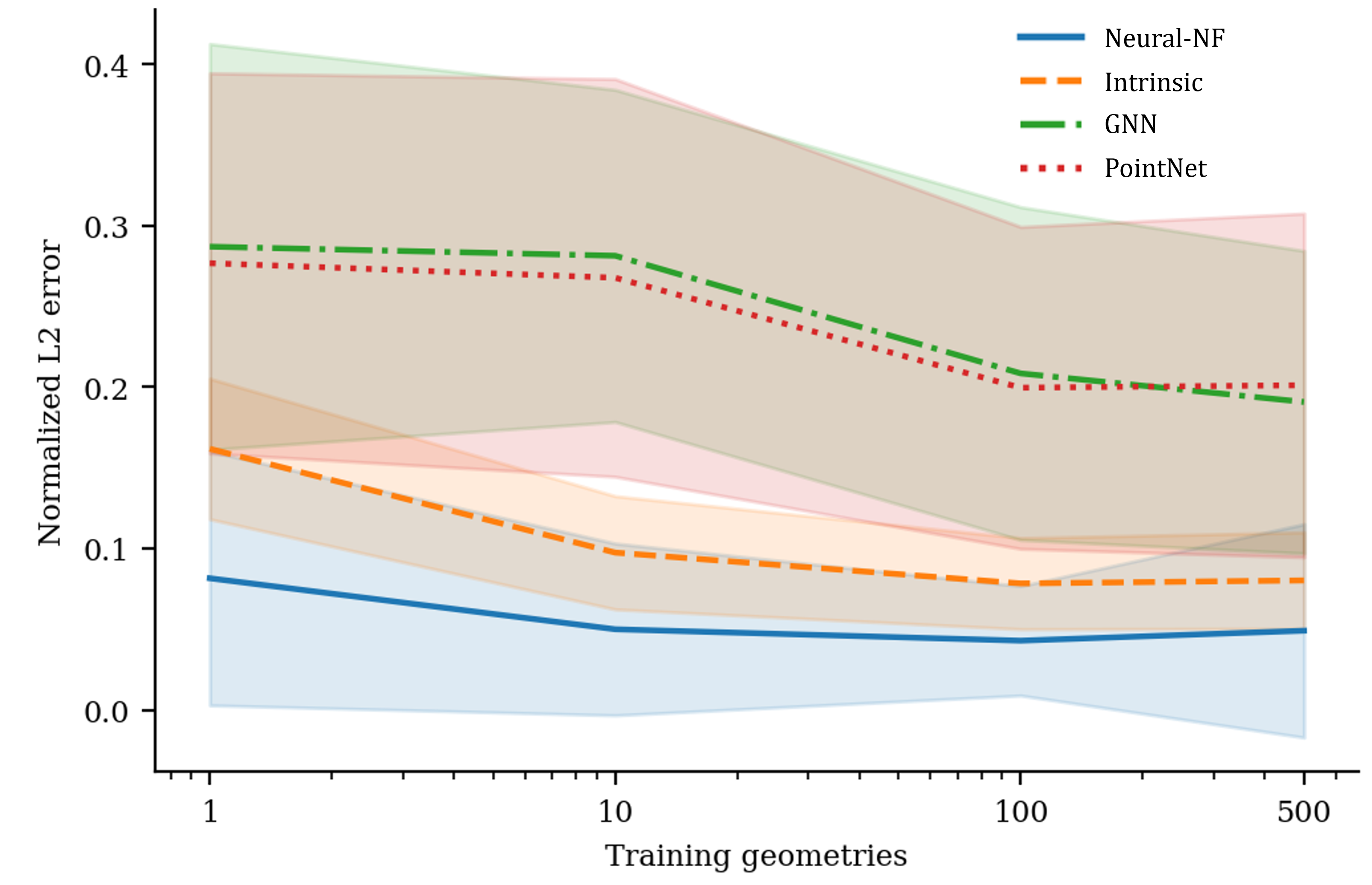}
    \caption{We demonstrate accurate navigation policies based on only a single training geometry (evaluated over many test geometries and goal locations), while baselines require more examples to provide similar performance. This was evaluated over the \emph{HouseExpo} example.}\label{fig:data_eff}
\end{wrapfigure}
Operator-learning planners such as \citep{matada2024generalizable} use regular-grid representations, so we compare against mesh-native direct predictors (GNN, GNO, MGN) \citep{pfaff2020learning,li2020neural,khan2020graph}. Neural-NF uses a feedforward coefficient network trained through sparse linear solves. Implementation details are provided in Appendix~E.
The eikonal objective is used as a standard reference task for geometry-dependent navigation, where the target value function \(V_{\mathrm{eik}}\) is the geodesic distance-to-goal solving
\(
|\nabla V_{\mathrm{eik}}(x)| = 1 \; \text{in } \Omega_g\setminus G,
\;
V_{\mathrm{eik}}=0 \; \text{on } G .
\)
To demonstrate the adaptability benefit of learning the operator we also evaluate non-eikonal objectives with clearance cost terms in Appendix~\ref{app:adaptable_obj}.

We evaluate Neural-NF on five 2D navigation datasets of increasing geometric complexity: \emph{Square}, \emph{Disk}, \emph{Maze}, \emph{HouseExpo}~\citep{li2020houseexpo}, and \emph{City Streets}~\citep{sturtevant2012benchmarks}. For each dataset we report value function error on held-out geometries from the same distribution (ID) and from a qualitatively different geometry class (OOD). For the synthetic examples the OOD is controlled by specifying the distribution of obstacles or maze complexity; we define the OOD for the HouseExpo as the City dataset, and respectively for the City Streets models. As shown in Table~\ref{tab:main_table}, Neural-NF substantially outperforms all direct-prediction models with Euclidean inputs in both settings. The improvement is most pronounced on complex geometries such as the \emph{Maze} dataset where baselines plateau near 40\% error regardless of architecture, while Neural-NF achieves 3.7\% ID and 7.4\% OOD, a roughly $5\times$ reduction. The intrinsic features alone greatly improve performance on more complicated geometries, due to directly encoding the topology of the domain. However, as shown in Figure~\ref{fig:sample_paths} and Table~\ref{tab:nav_results}, the PDE based implementation provides much more regular and navigable potentials due to the global consistency of the solve. The baseline error degrades under OOD shift across all datasets, whereas Neural-NF error remains relatively constant, supporting the hypothesized generalization.
Table~\ref{tab:nav_results} (right) further shows that this advantage extends to untrained metrics: in the \emph{Maze} OOD setting, Neural-NF achieves a navigation policy cosine error of $0.15 \pm 0.05$ compared to baselines ranging from $0.70$-$0.93$, and produces smoother gradient fields which is likely a result of the regularity of the elliptic solve, and may be beneficial for navigation. We additionally evaluate a wall-avoiding corridor objective with reduced speed near obstacles. Neural-NF retains strong OOD transfer under this adapted objective, achieving the lowest error across all three evaluated geometry families, as shown in Appendix~\ref{app:adaptable_obj}.

The ablation in Table~\ref{tab:logplan_ablation} isolates the contribution of both the Laplacian derived features and global PDE structure.  We compare the full Neural-NF implementation to an unlearned PDE with fixed coefficients and a direct prediction model using the intrinsic features.
The Intrinsic model improves value reconstruction but not policy accuracy, whereas the unlearned PDE provides a better policy but poor value accuracy. Neural-NF combines both advantages, reducing policy error from $0.19\pm0.09$ to $0.02\pm0.02$, thereby highlighting the role of both the local learning and global assembly.

Neural-NF also exhibits strong data efficiency. As shown in Figure~\ref{fig:data_eff}, our method produces accurate navigation policies from a single training geometry on the \emph{HouseExpo} dataset, while baselines require more examples to provide comparable performance and exhibit higher variance throughout. Figure~\ref{fig:single_shot_sweeps} extends this to single-shot extrapolation under both geometric perturbation and goal displacement, where Neural-NF maintains low error across the full sweep.
Finally, we provide a physical demonstration on the Crazyflie 2.0 platform, applying Neural-NF to pre-specified OOD environments shown in Figure~\ref{fig:fig1}, see Appendix~\ref{app:physical_experiments} for details.

\begin{table*}[t]
\centering
\small
\begin{minipage}{0.45\textwidth}
\small
\centering
\caption{We compare the ablations of the proposed method with \emph{i.} the PDE formulation alone and no learnable component and \emph{ii.} a learned model on the local features alone with no PDE formulation for the \emph{HouseExpo} example.}
\vspace{1.3cm}
\label{tab:logplan_ablation}
\begin{tabular}{lrrrr}
\toprule
 Method & $|| V-V_{\textrm{eik}}||_2$ & $d_{\cos}(u,u_{\textrm{eik}})$ \\
\midrule
Unlearned PDE & 0.38 ± 0.11 & 0.15 ± 0.05 \\
Intrinsic, no PDE & 0.07 ± 0.05 & 0.19 ± 0.09 \\
\midrule
\textbf{Neural-NF} & \textbf{0.05} ± 0.04 & \textbf{0.02} ± 0.02 \\
(intrinsic + PDE) & & \\
\bottomrule
\end{tabular}
\end{minipage}
\hfill
\begin{minipage}{0.45\textwidth}
\small
\centering
\caption{We evaluate navigation-relevant metrics in the challenging Maze OOD setting, as an extension of Table~\ref{tab:main_table}. We consider the cosine distance on the resulting policy and the smoothness, neither are explicitly trained but are best for the Neural-NF method.}
\label{tab:nav_results}
% \vspace{0.27cm}
\begin{tabular}{lrr}
\toprule
Method & $d_{\cos}(u,u_{\rm eik})$ & $||\sigma_{\nabla u}||$ \\
\midrule
PointNet & 0.70 $\pm$ 0.08 & 0.97 $\pm$ 0.19 \\
DeepONet & 0.71 $\pm$ 0.06 & 1.37 $\pm$ 0.18 \\
GNN & 0.75 $\pm$ 0.04 & 1.47 $\pm$ 0.18 \\
GNO & 0.92 $\pm$ 0.02 & \cellcolor{red!10} 0.71 $\pm$ 0.04 \\
MGN & 0.93 $\pm$ 0.01 & 0.88 $\pm$ 0.04 \\
Intrinsic & \cellcolor{red!10}  0.46 $\pm$ 0.18 & 1.13 $\pm$ 0.25 \\
\midrule
\textbf{Neural-NF} & \cellcolor{blue!5} \textbf{0.15} $\pm$ 0.05 & \cellcolor{blue!5} \textbf{0.53} $\pm$ 0.08 \\

\bottomrule
\end{tabular}
\end{minipage}

\end{table*}

\paragraph{Conclusions}
\label{sec:conclusions}
We introduced Neural-NF, a local-to-global framework for geometry-generalizable feedback motion planning. Rather than directly predicting a policy field, the method learns local geometry-dependent coefficients of an elliptic planning problem and solves the induced problem on each target domain. This lets the planner adapt to data-driven objectives while retaining the structure of value function-based planning.
The construction combines intrinsic Laplacian features, learned operator coefficients, and a global PDE solve to produce a consistent navigation function. It also provides an optimal-control interpretation, obstacle boundary-tangent velocities, monotone value function descent, and geometric inductive biases such as equivariance and local stability.
Empirically, the method improves data efficiency and transfer across unseen geometries compared with direct learned planners. The results support the central claim that learning the planning problem, as opposed to only the policy field, is a useful direction for robust feedback motion planning.

\paragraph{Limitations}
\label{sec:limitations}
Neural-NF poses learning over the structured class of navigation value functions represented by the proposed elliptic planner, rather than arbitrary policies or value fields. This structure provides the planner properties central to our method, but may not capture objectives or robot dynamics that require a different PDE or control formulation. We provide a full discussion of limitations and future work in Appendix~\ref{app:limitations}.

% All submissions should include a Limitations section (counted toward the 8-page limit), explicitly describing limiting assumptions, failure modes, and other limitations of the results and experiments, and how these might be addressed in the future.
%===============================================================================

% \clearpage
% The acknowledgments are automatically included only in the final and preprint versions of the paper.
\acknowledgments{M. A. Hsieh's work is supported by NSF Awward 2121887. B. Shaffer's work is supported by the National Science Foundation Graduate Research Fellowship under Grant No. DGE-2236662. N. Trask's work is supported by SEA-CROGS (Scalable, Efficient and Accelerated Causal Reasoning Operators, Graphs and Spikes for Earth and Embedded Systems), a Mathematical Multifaceted Integrated Capability Center (MMICCs) and the Department of Energy early career program.}

%===============================================================================

% \small
\bibliography{bib}  % .bib
% \normalfont
\appendix

\section{Optimal control correspondence}
\label{app:lsoc}

For completeness, we summarize the correspondence between the learned desirability PDE and linearly-solvable optimal control (LSOC)~\cite{kappen2005linear, todorov2009compositionality} for the Dirichlet-enforced goal formulation. For any admissible coefficients \(K_\theta\succ 0\) and \(c_\theta\ge 0\), the policy \eqref{eq:desire_transform} can be interpreted as the optimal stationary feedback of a stochastic control problem. Consider the reflected controlled diffusion on \(\Omega_g\),
\[
    dX_t = \bigl(a_\theta(X_t)+u_t\bigr)\,dt + \Sigma_\theta(X_t)\,dW_t - n(X_t)\,dL_t, \qquad \Sigma_\theta\Sigma_\theta^\top = K_\theta,
\]
with first hitting time \(\tau_G\) of the goal set where $n$ is the outward normal and $L$ is the boundary local time. The last term is the Skorokhod reflection term which pushes trajectories into the domain, and corresponds to the Neumann BCs used in the PDE formulation. The cost is
\[
J^\pi(x)=\mathbb{E}_x^\pi\!\left[ \int_0^{\tau_G} \left( q_\theta(X_t)+\tfrac12 u_t^\top K_\theta(X_t)^{-1}u_t \right)dt+ g(X_{\tau_G}) \right],
\]
where $q$ is the state-dependent running cost, $K_\theta^{-1}$ is the control cost weighting, and $g$ is the terminal cost at the goal. The LSOC linearization depends on the relation between the control cost and diffusion which results in the quadratic cancellation later, this is a structural choice in our interpretation of the learned model.
The associated stationary HJB equation is
\[
    0=\min_u\left\{ q_\theta+\nabla V_\theta^\top(a_\theta+u) +\tfrac12\operatorname{tr}(K_\theta\nabla^2 V_\theta) +\tfrac12 u^\top K_\theta^{-1}u \right\}.
\]
Minimizing over \(u\) gives the optimal feedback
\[
    u^*(x)=-K_\theta(x)\nabla V_\theta(x),
\]
and substituting this feedback into the HJB gives
\[
    0=q_\theta+a_\theta\cdot\nabla V_\theta
    +\tfrac12\operatorname{tr}(K_\theta\nabla^2 V_\theta)
    -\tfrac12\nabla V_\theta^\top K_\theta\nabla V_\theta .
\]

The key LSOC step is the logarithmic desirability transform \(z_\theta=e^{-V_\theta}\), in PDEs this linearization is known as the Cole-Hopf transformation. Since
\[
    \nabla V_\theta=-\frac{\nabla z_\theta}{z_\theta},
    \qquad
    \nabla^2V_\theta
    =
    -\frac{\nabla^2z_\theta}{z_\theta}
    +
    \frac{\nabla z_\theta\nabla z_\theta^\top}{z_\theta^2},
\]
the quadratic gradient terms cancel (see \citep{kappen2005linear}). With $z_\theta>0$ this yields the linear equation
\[
    a_\theta\cdot\nabla z_\theta +\tfrac12\operatorname{tr}(K_\theta\nabla^2 z_\theta) -q_\theta z_\theta=0.
\]
We particularly choose the passive drift \(a_\theta=\tfrac12\nabla\cdot K_\theta\), corresponding to the reversible diffusion associated with \(K_\theta\) to provide the generator
\[
    \mathcal L_\theta f = a_\theta\cdot\nabla f +\tfrac12\operatorname{tr}(K_\theta\nabla^2 f) = \tfrac12\nabla\cdot(K_\theta\nabla f),
\]
and the desirability equation becomes
\[
    \nabla\cdot(K_\theta\nabla z_\theta)-c_\theta z_\theta=0,
    \qquad c_\theta=2q_\theta.
\]
With the reflecting (in the path sense) boundary condition \((K_\theta\nabla z_\theta)\cdot n=0\) on \(\partial\Omega_g\) and absorbing goal condition \(z_\theta|_G=e^{-g}\), this is the homogeneous LSOC form corresponding to the learned elliptic planner. Finally, because \(V_\theta=-\log z_\theta\),
\[
    u^*(x)=-K_\theta\nabla V_\theta
    =
    K_\theta\nabla\log z_\theta,
\]
which matches \eqref{eq:desire_transform}.

Thus \(K_\theta\) simultaneously defines the diffusion tensor, the control-cost metric, and the PDE conductivity, while \(c_\theta\) defines the running cost. Learning over \((K_\theta,c_\theta)\) can therefore be interpreted as learning a structured family of stochastic navigation objectives. The exact LSOC correspondence holds for the absorbing-goal Dirichlet formulation; in the experiments, we use the smooth goal-forced variant in \eqref{eq:desire_pde}.

\section{Geometric interpretation}
\label{app:geometric_interpretation}
By the logarithmic transform, the model represents value functions satisfying the viscous Hamilton-Jacobi equation (correspondence in Appendix~\ref{app:lsoc})
\[
    \nabla V_\theta^\top K_\theta \nabla V_\theta - \nabla\cdot(K_\theta \nabla V_\theta) = c_\theta.
\]
The first term corresponds to anisotropic travel-time (eikonal) dynamics, while the second term provides viscosity regularization. The underlying first-order equation
\[
    \nabla V^\top K_\theta \nabla V = c_\theta
\]
defines geodesic distance under the Riemannian metric $g_\theta(x) = c_\theta(x)\,K_\theta(x)^{-1}$. Thus, the proposed model can be interpreted as learning a viscosity-regularized geodesic planner under a geometry-dependent metric, which provides a useful inductive bias for navigation tasks \citep{viswanath2026physics}.

\section{Intrinsic feature construction}
\label{app:intrinsic_features}
\begin{wrapfigure}{r}{0.35\textwidth}
    \vspace{-\intextsep}
    \centering
    \includegraphics[width=\linewidth]{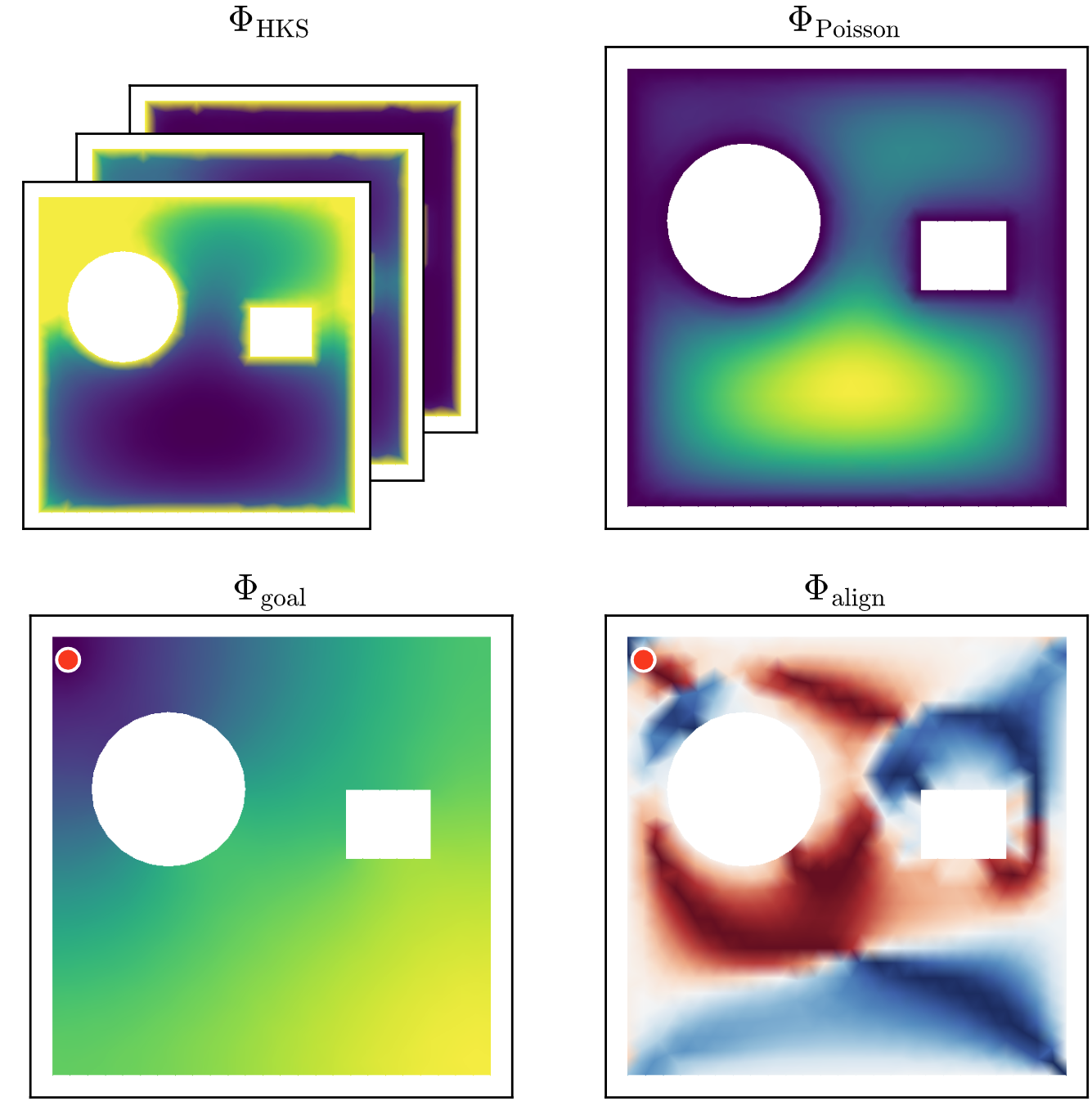}
    \caption{Intrinsic features for a given geometry, all are constructed from the Laplacian on the mesh and naturally invariant in rigid model.}
    \label{fig:intrinsic_features}
\end{wrapfigure}
We construct intrinsic feature channels as functions of the Laplace operator $\mathcal{L}_g$ on a mesh discretizing $\Omega_g$. We briefly define each of these features, show an example realization in Figure~\ref{fig:intrinsic_features}, and provide a simple ablation on them in Table~\ref{tab:logplan_ablation}.

% \textbf{HKS:}
The HKS channels provide a multi-scale descriptor of local geometry which is widely used in computer graphics and can be constructed either from discrete time simulation or via the diagonalized Laplacian if available,
\[\Phi_{\mathrm{HKS}}(i,j) = (e^{-t_j \mathcal{L}_g}\delta_{x_i})(x_i) \in \mathbb{R}^{N\times T},\]
we use $T=8$ for all experiments.

The wall-distance is naturally encoded through the Laplacian via Poisson equation with Dirichlet boundary conditions and uniform forcing, 
solve $-\Delta u = 1$ in $\Omega_g$, $u|_{\partial\Omega_g}=0$, and define 
\[\Phi_{\mathrm{Poisson}}(i)=u(x_i)\in\mathbb{R}^{N\times1}.\]

We provide a smooth embedding of the goal within the domain by solving 
$-\Delta u_G + c_{\rm min}u_G= \rho_G$ in $\Omega_g$, with Neumann BCs, for a small constant $c_{\rm min}$ and then define the feature as
\[\Phi_{\mathrm{goal}}(i)=u_G(x_i)\in\mathbb{R}^{N\times1}.\]

We define a source-dependent directional signal by comparing the goal encoding with the geometry-fixed Poisson depth field as,
\[\Phi_{\mathrm{align}}(i)=\frac{\nabla \Phi_{\mathrm{goal}}(x_i)\cdot\nabla \Phi_{\mathrm{Poisson}}(x_i)}{\|\nabla \Phi_{\mathrm{goal}}(x_i)\|\,\|\nabla \Phi_{\mathrm{Poisson}}(x_i)\|+\varepsilon} \in\mathbb{R}^{N\times 1}.\]

\section{Equivariance}
\label{app:equivariance}

We show that the Neural-NF policy satisfies $u_\theta(Rx;\,Rg,\,RG)=R\,u_\theta(x;\,g,\,G)$ for any rigid motion $R$, by tracing the pipeline in \eqref{eq:pipeline}.
First, the Laplace operator is intrinsic and commutes with rigid motions as $\mathcal{L}_g(f\circ R) = (\mathcal{L}_{Rg}f)\circ R$ \citep{canzani2013analysis}. Since every feature channel is constructed from $\mathcal{L}_g$ and the goal $G$, each is a scalar invariant.
Because $\mathcal{Q}_\theta$ acts pointwise on these scalar features, the coefficient fields inherit the same invariance.
The desirability PDE on $\Omega_{Rg}$ is then identical in form to that on $\Omega_g$, so by uniqueness $z_\theta(Rx;Rg,RG)=z_\theta(x;g,G)$.

Finally, $u_\theta = K_\theta\nabla\log z_\theta$. For orthogonal $R$, gradients transform as $\nabla_{Rx} = R^{-\top}\nabla_x = R\nabla_x$, so
\begin{align*}
    u_\theta(Rx;\,Rg,\,RG) 
&= k_\theta(x)\,\nabla_{Rx}\log z_\theta(Rx)\\
&= k_\theta(x)\,R^{-\top}\nabla_x\log z_\theta(x)\\
&= k_\theta(x)\,R\,\nabla_x\log z_\theta(x)\\
&= R\,k_\theta(x)\nabla_x\log z_\theta(x)\\
&= R\,u_\theta(x;\,g,\,G).
\end{align*}
The property holds for every $\theta$ by construction and removes the need for any data augmentation in this regard.

\section{Experiment details}
\label{app:experimental_details}

For all models we optimize using Adam with fixed learning rate, $1e-4$ for $10k$ training steps sampled randomly over the training data with fixed batch size of $8$. We use a fixed hidden dimension of 128 and 4 hidden layers for our Neural-NF networks, and similar hyperparameters for baselines where applicable. We batch over the linear PDE solve for Neural-NF via multi-threading on CPU. All baselines are provided spatial coordinates and Signed Distance Function (SDF) inputs, along with the goal location in spatial coordinates, consistent with previous work. Other methods have included occupancy inputs which are not needed in the structured setting where all points are within the domain and boundary. The Intrinsic baseline is constructed from the intrinsic input features with a neural-field directly outputting the navigation function, making this a direct ablation on the PDE solve in the Neural-NF approach. Mass matrices were assembled in Scikit-FEM \citep{skfem2020}.

\subsection{Synthetic data generation}
\label{app:data_gen}
\begin{figure}
    \centering
    \includegraphics[width=0.5\linewidth]{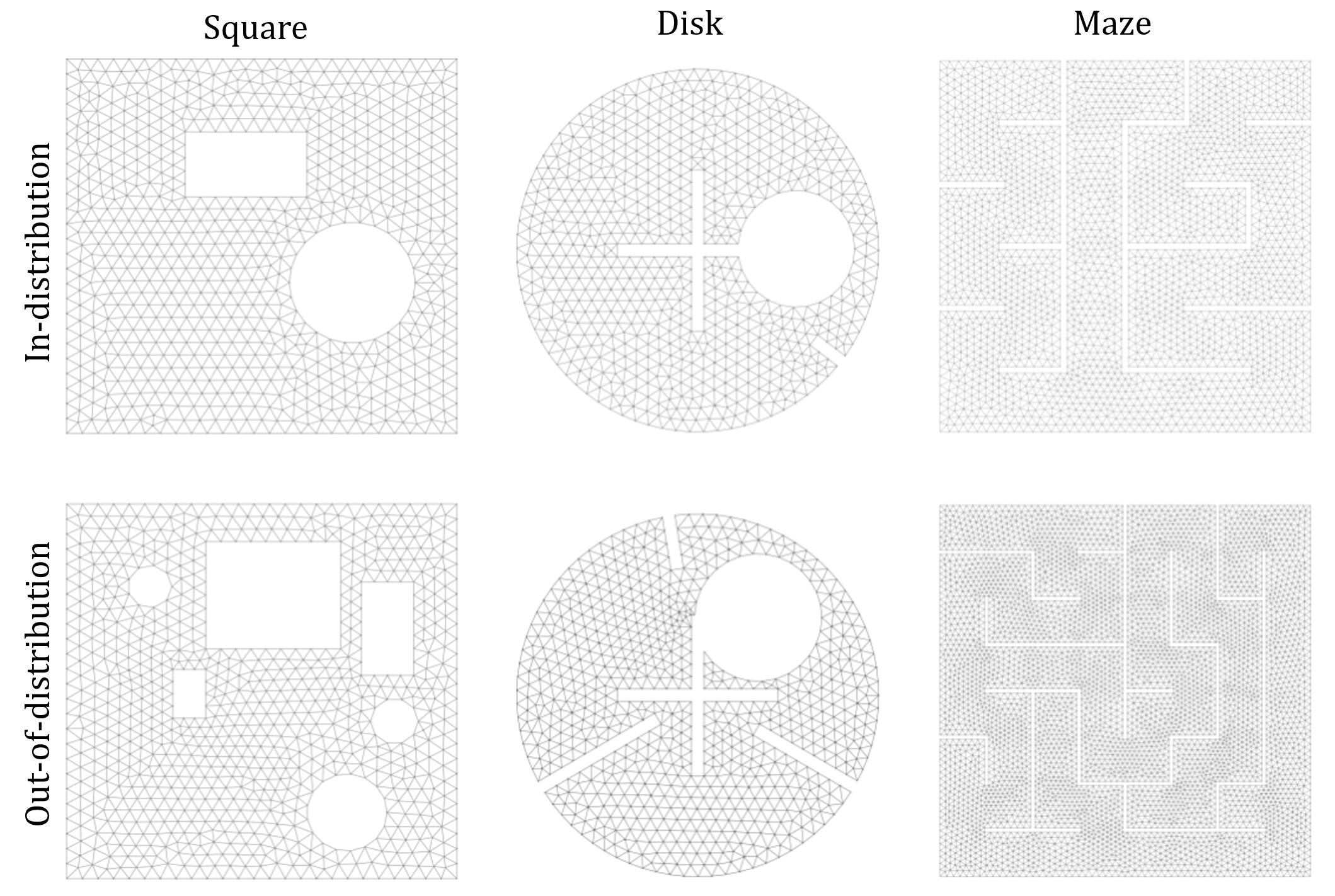}
    \caption{Synthetic data cases used in the main results showing a single ID and OOD sample for each case. As shown in Table~\ref{tab:main_table}, these correspond to approximately increasing difficulty as measured by the baseline method error.}
    \label{fig:synthetic_data}
\end{figure}
We generated our synthetic example cases through pyGmsh by algorithmically specifying the distributions of geometries and obstacles. We use different distributions (obstacle size, number, or maze resolution) to control the ID vs OOD cases. Examples of ID and OOD geometries for each experiment are shown in Figure~\ref{fig:synthetic_data}. On the synthetic geometries, we use the fast marching method \citep{sethian1999fast} with variable speeds on the unstructured meshes to generate target values for supervised training.

\section{Additional results}
\label{app:add_results}

\begin{figure}
    \centering
    \includegraphics[width=0.7\linewidth]{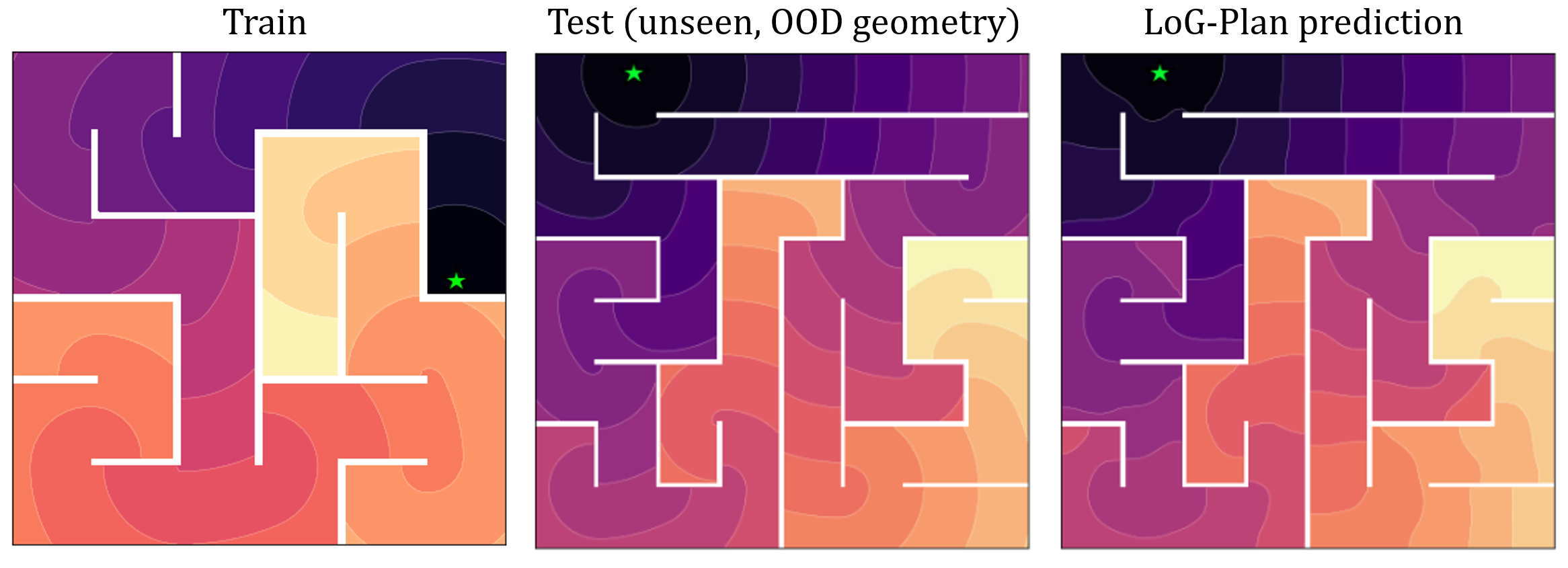}
    \caption{Example target functions on training and OOD test geometry for the maze task, the Neural-NF model provides an accurate and smooth prediction on a more complex and unseen geometry despite low training data of only 100 configurations. Goal locations are shown in green.}
    \label{fig:maze_demo}
\end{figure}

We provide an additional visualization of the training and test geometries for the challenging maze example with Neural-NF for comparison. Over unseen and higher complexity maze configurations our approach maintains consistent low error, as a result of the guarantees of the navigation function formulation and the strong inductive biases for learning. This is shown in Figure~\ref{fig:maze_demo}.

\subsection{Adaptable navigation objective}
\label{app:adaptable_obj}

\begin{figure}
    \centering
    \includegraphics[width=0.5\linewidth]{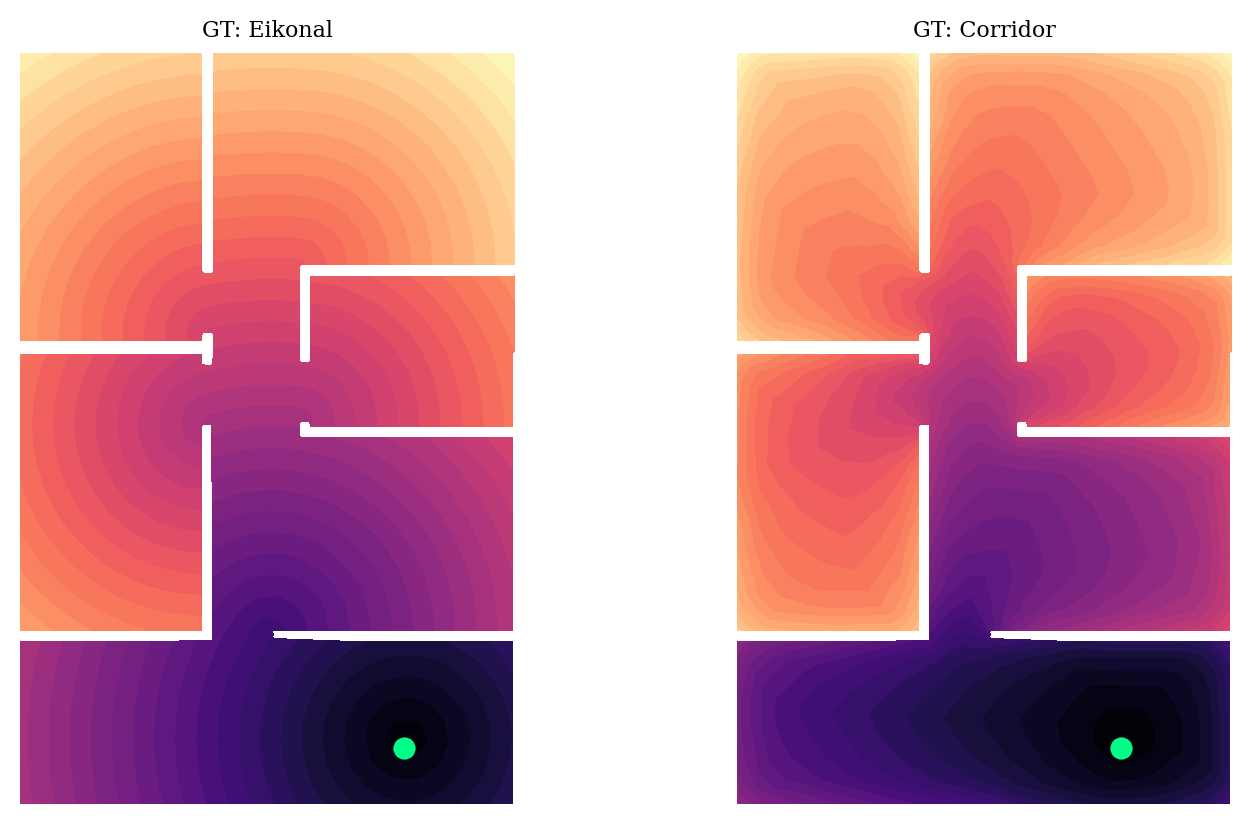}
    \caption{Target functions for the eikonal and wall-avoiding (corridor) objectives.}
    \label{fig:exp2_gt_comparison}
\end{figure}

The main benefit of the Neural-NF approach over conventional planning methods is the ability to leverage automatic differentiation to train the defined planner formulation through gradient descent over the intrinsic geometric features. In principle this could be used with any planning related objective function, in the main text we consider minimal distance objectives as those are most often considered in path-planning tasks. In more complex motion planning tasks we typically prefer navigation policies which, for instance, provide faster routes under a specified platform dynamics. To demonstrate our ability to support this class of objectives, we consider a supervising navigation function for a policy with slower motion near walls as a representative example. The ground truth policies for the eikonal problem class and the \emph{corridor} objective are shown in Figure~\ref{fig:exp2_gt_comparison}. We report the results over the synthetic datasets in Table~\ref{tab:appendix_additional_l2}. Similarly to the results in the main text, Neural-NF provides a substantial improvement in relative $L_2$ on the navigation function compared to baseline learning approaches.

\begin{figure}
    \centering
    \includegraphics[width=1\linewidth]{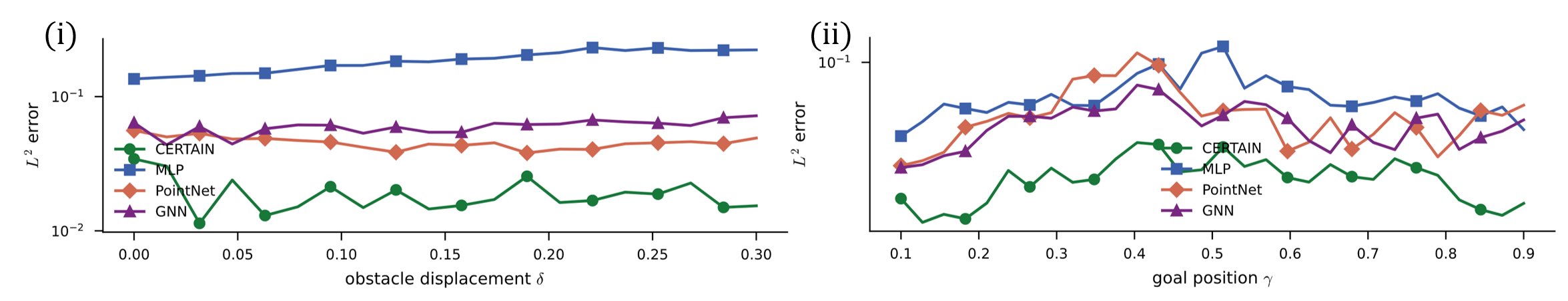}
    \caption{Our approach enables \emph{single shot extrapolation} for (i) geometric perturbation, and (ii) goal location, showing improved predictions in both cases over baselines, trained from only a single sample. } 
    \label{fig:single_shot_sweeps}
\end{figure}

\begin{table}[h]
\small
\centering
\caption{Comparison of $L_2$ error on the learned value function for the \emph{corridor} objective function setting on three representative test geometry families. We evaluate on both unseen test geometries drawn from the same distribution as training (ID) and geometries drawn from a different distribution (OOD). The results are overall consistent as for the eikonal objective.}
\label{tab:appendix_additional_l2}
\begin{tabular}{lrrr}
\toprule
\textbf{ID} $\| V-V_{\textrm{target}}\|_2\, (\%)$ 
    & Square & Disk & Maze \\
\midrule
PointNet
    & \cellcolor{red!10} 7.9 $\pm$ 2.8
    & 22.9 $\pm$ 11.3
    & 40.8 $\pm$ 9.8 \\
GNN
    & 8.9 $\pm$ 3.8
    & 24.1 $\pm$ 11.5
    & 38.8 $\pm$ 9.7 \\
DeepONet
    & 13.9 $\pm$ 4.9
    & 31.8 $\pm$ 14.7
    & 40.8 $\pm$ 9.3 \\
GNO
    & 9.1 $\pm$ 3.7
    & 25.1 $\pm$ 12.0
    & 38.2 $\pm$ 10.0 \\
MGN
    & 8.7 $\pm$ 3.6
    & 21.3 $\pm$ 9.9
    & 39.5 $\pm$ 10.1 \\
Intrinsic
    & \cellcolor{red!10} 7.9 $\pm$ 3.2
    & \cellcolor{red!10} 10.8 $\pm$ 6.6
    & \cellcolor{blue!5} \textbf{2.7} $\pm$ 1.3 \\
\textbf{Neural-NF}
    & \cellcolor{blue!5} \textbf{4.0} $\pm$ 1.8
    & \cellcolor{blue!5} \textbf{5.5} $\pm$ 4.1
    & \cellcolor{red!10} 3.5 $\pm$ 3.7 \\
\midrule
\textbf{OOD} $\| V-V_{\textrm{target}}\|_2\, (\%)$ 
    & Square & Disk & Maze \\
\midrule
PointNet
    & 11.7 $\pm$ 5.6
    & 29.1 $\pm$ 14.8
    & 44.1 $\pm$ 9.2 \\
GNN
    & 13.2 $\pm$ 6.2
    & 26.7 $\pm$ 13.1
    & 41.4 $\pm$ 8.7 \\
DeepONet
    & 20.9 $\pm$ 10.1
    & 35.2 $\pm$ 17.6
    & 42.4 $\pm$ 8.4 \\
GNO
    & 13.5 $\pm$ 6.2
    & 26.4 $\pm$ 12.7
    & 42.2 $\pm$ 8.7 \\
MGN
    & 13.3 $\pm$ 6.3
    & 24.9 $\pm$ 12.1
    & 44.4 $\pm$ 8.6 \\
Intrinsic
    & \cellcolor{red!10} 10.6 $\pm$ 5.5
    & \cellcolor{red!10} 13.4 $\pm$ 10.2
    & \cellcolor{red!10} 14.1 $\pm$ 8.9 \\
\textbf{Neural-NF}
    & \cellcolor{blue!5} \textbf{7.8} $\pm$ 5.9
    & \cellcolor{blue!5} \textbf{7.0} $\pm$ 5.2
    & \cellcolor{blue!5} \textbf{8.9} $\pm$ 5.4 \\
\bottomrule
\end{tabular}
\end{table}

\subsection{Single-shot extrapolation}
We evaluate the ability of our proposed Neural-NF framework to extrapolate from a single training instance, both with respect to varying goal location and obstacle location for a single obstacle. In both cases we maintain $<5\%$ error over the range of values and outperform the direct prediction approaches over all values.

\subsection{Ablation on input modalities}
We present a general framework for constructing geometrically stable inputs built on the mesh Laplacian. There are many possible choices for input features constructed from this that satisfy the Lipschitz constraint. We present ablations on the $L_2$ and $cos$ distance errors from models trained in a uniform setting over varying inputs.

\subsection{Physical Experiments}
\label{app:physical_experiments}
\begin{wrapfigure}{r}{0.4\textwidth}
    \vspace{-\intextsep}
    \centering
    \includegraphics[width=\linewidth]{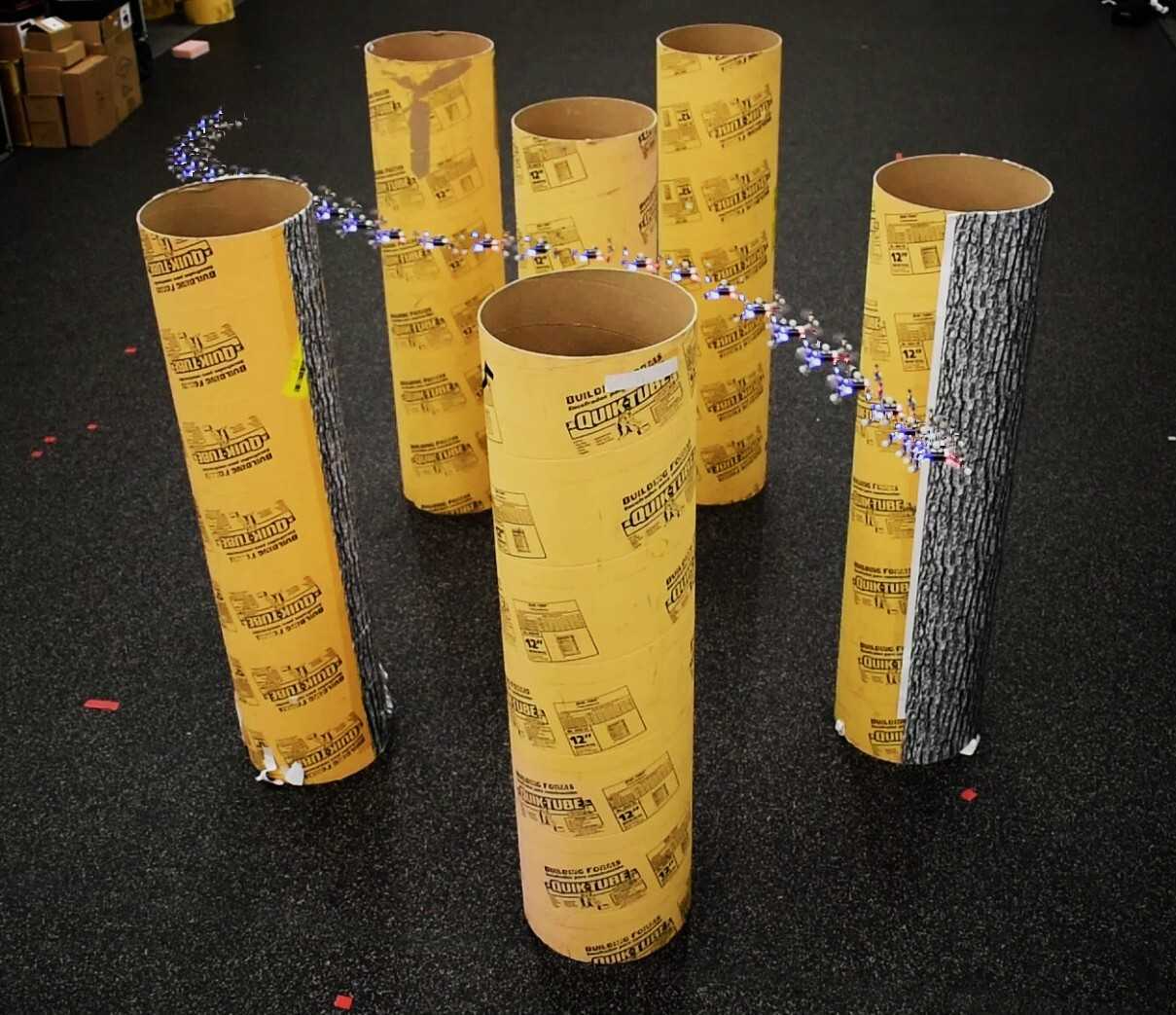}
    \caption{Physical experiment time-lapse of a Crazyflie 2.0 quadrotor navigating through \textbf{Case 1} while following a collision-free reference trajectory.}
    \label{fig:physical_timelapse}
\end{wrapfigure}
We validate our Neural-NF framework in physical experiments using a single Bitcraze Crazyflie 2.0 quadrotor. A laptop equipped with an Intel i5 CPU serves as the base station, receiving position and orientation estimates from a Vicon motion capture system at 120 Hz. The base station computes thrust and angular velocity commands at 100 Hz using a feedback linearization controller to track the navigation policy at a fixed height of 1 m, and sends the commands to the quadrotor through a Crazyradio PA dongle.

The physical feasibility and generalizability of our approach are evaluated on two OOD environments, as shown in Figure~\ref{fig:fig1}. \textbf{Case 1} contains multiple obstacles, while \textbf{Case 2} includes geometric features qualitatively different from those in the training dataset, which consists only of a single circular and a single rectangular obstacle. In both cases, the quadrotor successfully navigates to the goal without colliding with any obstacles. A time-lapse of the quadrotor navigating through \textbf{Case 1}, where the obstacles are formed by cylinders with a diameter of 12 inches, is shown in Figure~\ref{fig:physical_timelapse}.

These experiments demonstrate the practical advantages of our elliptic PDE approach, which generates safe and convergent navigation policies due to the built-in regularity of the navigation value function with respect to geometry and goal locations. Future work should extend this demonstration to a full closed-loop sense-and-plan schema, which would require generating the input geometries from, e.g., onboard LiDAR data.

\section{Full limitations}
\label{app:limitations}
Neural-NF is designed for value functions that arise from the proposed elliptic planning formulation, as opposed to arbitrary navigation value functions. This restriction is intentional as the PDE imposes regularity, boundary behavior, and monotone descent properties, but it also limits the hypothesis class. Other applications may require different boundary conditions, coefficient parameterizations, or extensions of the underlying optimal-control model~\citep{todorov2009compositionality,dvijotham2012linearly}.
The method also depends on the expressivity of the intrinsic feature map. Our Laplacian-derived features work well empirically, but richer local or task-dependent descriptors may be needed for more complex robot dynamics or environments. In the main text we focus on 2D navigation, although the formulation extends directly to 3D.
Finally, our experiments use supervised targets, while the same differentiable solve could support residual-based, rollout-based, or reinforcement-learning objectives in future work \citep{matada2024generalizable}.

\begin{table}[]
    \centering
    \caption{Ablation on input modalities for the Neural-NF method.}
    \begin{tabular}{lrr}
    \toprule
     & L2 error (↓) & cos error (↓) \\
    \midrule
    Depth & 0.151 ± 0.061 & 0.063 ± 0.021 \\
    Depth + HKS & 0.137 ± 0.067 & 0.046 ± 0.024 \\
    Neural-NF (Full) & \textbf{0.071} ± 0.069 & \textbf{0.030} ± 0.027 \\
    \bottomrule
    
    \end{tabular}
\end{table}

\end{document}